\documentclass[a4paper]{article}
\usepackage[preprint]{spconf}
\usepackage{amsmath,graphicx}
\usepackage{listings}
\usepackage{caption}
\usepackage{subcaption}
\usepackage{amssymb}
\usepackage{amsmath}
\usepackage{multirow}
\usepackage{xcolor}
\usepackage{float}
\usepackage{lipsum}
\usepackage{hyperref}
\usepackage{array}

\usepackage[absolute,overlay]{textpos}

\newcolumntype{A}{>{\scriptsize } l}
\usepackage[capitalize]{cleveref}
\Crefname{chapter}{Chap.}{ChapTs.}
\Crefname{section}{Sec.}{Secs.}
\Crefname{figure}{Fig.}{Figs.}

\title{Improved robustness to disfluencies in RNN-Transducer based Speech Recognition}

\name{Valentin Mendelev$^{\star 1}$, Tina Raissi$^{\star2\dagger}$, Guglielmo Camporese$^{3\dagger}$, Manuel Giollo$^1$ \thanks{$^\star$ Denotes equal contribution}\thanks{$\dagger$ Work done during an internship at Amazon} }
\address{$^1$ Amazon Alexa \\ 
$^2$ Human Language Technology and Pattern Recognition Group, RWTH Aachen University, Germany \\ 
$^3$ Department of Mathematics ``Tullio Levi-Civita", University of Padova, Italy \\
\{mendelev,mgiollo\}@amazon.com, raissi@rwth-aachen.de, guglielmo.camporese@phd.unipd.it}

\begin{document}
\begin{textblock*}{17cm}(2cm,27cm) 
\footnotesize{\copyright\ 2020 IEEE. Personal use of this material is permitted. Permission from IEEE must be obtained for all other uses, in any current or future media, including reprinting/republishing this material for advertising or promotional purposes, creating new collective works, for resale or redistribution to servers or lists, or reuse of any copyrighted component of this work in other works.}
\end{textblock*}

%
\maketitle
\begin{abstract} 

Automatic Speech Recognition (ASR) based on Recurrent Neural Network Transducers (RNN-T) is gaining interest in the speech community.\
We investigate data selection and preparation choices aiming for improved robustness of \mbox{RNN-T} ASR to speech disfluencies with a focus on partial words.\ For evaluation we use clean data, data with disfluencies and a separate dataset with speech affected by stuttering.\ We show that after including a small amount of data with disfluencies in the training set the recognition accuracy on the tests with disfluencies and stuttering improves.\ Increasing the amount of training data with disfluencies gives additional gains without degradation on the clean data.\ We also show that replacing partial words with a dedicated token helps to get even better accuracy on utterances with disfluencies and stutter.\ The evaluation of our best model shows 22.5\% and 16.4\% relative WER reduction on those two evaluation sets.

\end{abstract}


\section{Introduction}
\label{sec:intro}

Human speech typically contains disfluencies alongside the articulation of an intended word sequence. Speech from any speaker has filled pauses, partial words and repetitions, while certain speech disorders (e.g. stuttering) amplify these phenomena.\ And despite the general performance achievements in speech recognition using End-to-End~(E2E) models, there is still not enough robustness to them.\ The main objective of this work is to investigate training data filtering and transcription processing choices for an ASR system based on the Recurrent Neural Network Transducer~(RNN-T)\cite{rnnt}, which may improve its robustness to speech disfluencies with a special focus on the partial words and repetitions.\

Motivation for this work comes from discussions with our colleagues who reported that self corrections are responsible for significant share of entity resolution errors in several voice assistant use-cases.\ Also, we wanted to investigate if ASR robustness to disfluencies may be improved by overweighting the data with partial words in the training set and if this gives a higher ASR accuracy for speakers with stuttering, even though vast majority of those data came from fluent speakers.\

Main contributions of this work are: a study of the effect of different ways to represent partial words in transcripts used to train RNN-T system, experimental results with different fractions of data with disfluencies in the training set and a view of the influence of the factors mentioned before on ASR performance for speakers with stuttering.

In the next session an overview of the prior work is presented, then the datasets are described in \cref{sec:data}.\ The experimental setting and results are discussed in \cref{sec:experiments} which is followed by conclusions and future work.

\section{Prior work}
The initial attention of the research community towards speech disfluencies derives from the importance of the improvement of the ASR system accuracy not only for the speech signal which is recorded under controlled conditions but also for the spontaneous speech~\cite{preliminar,heeman1999speech}.\ The resulting task comprises the identification and the consecutive removal of the disfluency events in the recognizer output and is solved by using the \textit{noisy channel} approach~\cite{johnson2004tag,honal2003correction}.\ Following the Bayesian statistical framework, this entails maximization of the a-posteriori probability of the word sequence with disfluencies, given the originally intended word sequence.\  Since disfluency events affect different phonetic aspects of speech~\cite{shriberg1999phonetic} many researchers tried to take advantage of the possible combination of different sources of knowledge on both acoustic and language model sides~\cite{liu2005comparing,ferguson2015disfluency,zayats2016disfluency,alharbi2020sequence}.\ The disfluency detection task in all mentioned works is solved by using sequence labelling/tagging approaches which can rely either on a generative approach such as Hidden Markov Model or discriminative log-linear models, such as Maximum Entropy Markov Model or Conditional Random Fields as well as Bidirectional Long Short-Term Memory based networks in combination with an attention mechanism~\cite{bach2019noisy}.\ In most of the cases the overall system maintains its modular setting and therefore requires not only separate optimization criteria for different components but also in some cases hand-labeled features for the annotation of the disfluencies to train the language model.\ 
A recent work brings the focus on the acoustic side and does not take into consideration any language-dependent information~\cite{kourkounakis2020detecting}.\ To the best of the authors' knowledge, with the exception of a work on personalized ASR for dysarthric speech~\cite{shor2019personalizing}, none of the published papers are aimed to improve speech recognition accuracy of an E2E ASR system by dealing with disfluencies without solving disfluency detection task itself.

\section{Datasets}
\label{sec:data}


For our experiments we used subsets of the transcribed data pool available to train Alexa ASR models.\ The recordings comprising the data pool are anonimized voice assistant requests recorded with various far-field devices in compliance with terms of service.\ Each transcription, in addition to the spoken words, contains tags provided by the transcriber indicating additional information on the speech signal.\ The attribution of the described tags relies on the transcriber's perception and expertise and therefore can be source of possible inaccuracy for both tag and spoken word annotations.\ This aspect is especially valid when the utterance contains unintelligible or disfluent speech.\ Most disfluency events such as word or syllable prolongation are actually not marked in the transcriptions.\ 

In this work we use three datasets, which we call \textit{Ordinary}, \textit{Disfluencies} and \textit{Stutter}.\

The \textit{Ordinary} dataset contains ordinary utterances with intelligible device-directed speech but without partial words.\ Acoustic conditions may be challenging because of low signal-to-noise ratio, media speech or due to the presence of multiple speakers.\ 

The \textit{Disfluencies} dataset is derived by applying a set of filters, which operate on transcriptions level on the large pool of data.\ The filters aim to select challenging utterances with partial words, repetitions and hesitations.\ More specifically, an utterance is included into this dataset if its transcription contains a partial word and its subsequent completion (e.g.  'alarm on tw- on twelve') and at least one of the following conditions is true:  

- there are no more than 4 words in the transcription;

- there is at least one other partial word (not necessarily with completion);

- there are hesitations;

- there are repetitions.

\noindent
This set of filters was chosen after trying several alternatives and observing that without additional conditions the dataset contained a lot of utterances with a single partial word, which were considered not challenging enough.\ We have to note that after most of the experiments mentioned in this work were done we repeated some of them with the simplest filter, which accepted utterances with a single partial word in the transcript.\ We found that conclusions reported in the following sections were mostly valid for datasets derived with this simple filter as well.

The Ordinary and Disfluencies datasets include train, dev and test partitions, which do not have speaker overlap. 

The \textit{Stutter} dataset was recorded by a vendor and contains speech samples provided by 11 speakers with stuttering.\ The speakers were reading prompts containing possible requests to a voice assistant in a quiet acoustic environment. This dataset is used for the evaluation purpose only.

Size of the datasets is presented in \cref{tab:datastat}. We restricted the amount of training data in the Ordinary dataset to have faster turnaround time. 
Also, the data for the Disfluencies dataset were selected from an order of magnitude bigger data pool in comparison to the Ordinary Train to enable experiments with increased relative amount of challenging data in the training sets. 

 \begin{table}[H]		
	\centering
	\caption{Size of the datasets used in this work in hours of sound.}	
	\label{tab:datastat}
	\begin{tabular}{lcc} 		
		\hline
\textbf{Dataset} & \textbf{Train (hours)} & \textbf{Test (hours)} \\ \hline \hline
Ordinary & $\sim 2300$ & $>20$ \\
Disfluencies & $47$ & $5$ \\
Stutter & $\mbox{-}$ & $2$ \\ \hline
\end{tabular}
\end{table}

\subsection{Handling Partial Words in Transcriptions}

Once data with partial words are included in the training set, there are several options how to mark such words.\ In this work we assume that our goal is to have ASR output free from disfluencies. This can be achieved if the system: (1) ignores them, (2) outputs a label instead of the partial word, (3) concatenates a partial label with the disfluency content which later can be removed via the post-processing (e.g. for the recording with the reference transcript 'p- play' the system may output (1) 'play', (2) '$\langle\text{pw}\rangle$ play', (3) 'p$\langle\text{pw}\rangle$ play').\ We decided to test the 3 options mentioned plus the one where only the first letter of the partial word remains and is appended with $\langle\text{pw}\rangle$ on the right.\  The motivation behind the last two options is clear: ideally we would like to keep disfluency content in order to preserve more information for the downstream tasks. Still, the quality of the ASR output with disfluencies removed is considered as the main criterion in the current work.

\section{Experimental Setup and Results}
\label{sec:experiments}
\subsection{Experimental Setting}

We train models suitable for on-line recognition. The model consist of a 5 layers deep encoder, a 2 layers deep prediction network, a joint network as in \cite{graves2013speech}, and an output layer with a softmax nonlinearity.\ Each layer of the encoder and the prediction network comprises 1024 Long Short-Term Memory~\cite{hochreiter1997long} units. The size of the joint network layer is 512 and the output layer size is 4001 corresponding to 4000 wordpeices and a \textit{blank} symbol. The wordpiece model was trained on a large set of voice assistant requests using a unigram language model \cite{kudo2018sentencepiece}.

The model accepts 192 dimensional input feature vectors each comprising three 64 dimensional Log-Mel-Filterbanks extracted every 10 milliseconds and stacked together. 

Training objective is minimization of RNN-T loss function \cite{rnnt,graves2013speech} with Adam optimizer \cite{kingma2014adam}, with total batch size of 1536 utterances and warmup-hold-decay learning rate schedule. We also use SpecAugment \cite{park2019specaugment}.

Evaluations were done using a beam decoding with the beam size 16.\ Model specific post-processing was applied on both the hypothesis and the reference in order to get transcripts free from partial words. For models trained with replacement of the full partial word with $\langle\text{pw}\rangle$ tag, those were removed. For models where a partial word or its first letter persisted in the training transcript, the tag was removed together with all letters before first space to the left of it.

\subsection{Case Studies}
 
\begin{figure}[t]
	\centering
	\includegraphics[width=1\linewidth, trim={6cm 9cm 7.5cm 9.5cm},clip]{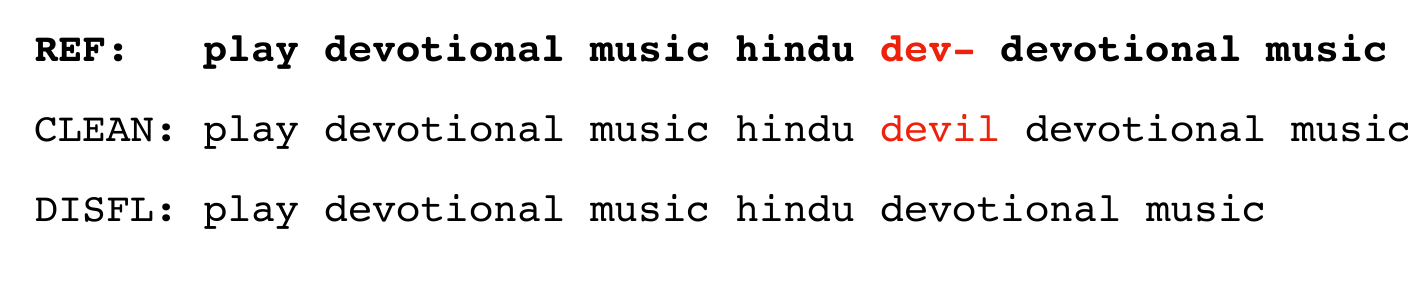}
	\caption{Example recognition results. \textit{REF} denotes the reference transcription, \textit{CLEAN} was produced by the baseline model, \textit{DISFL} -- by the model trained on Ordinary Train merged with Disfluencies Train and with partial words removed from training transcripts.}
	\label{fig:ex1}
\end{figure}

After training the baseline model on the Ordinary Train, the experimental models on the Ordinary Train merged with the Disfluencies Train datasets and different handling of partial words in the transcripts, we looked into the decoding results of Disfluencies Test in order to ensure that the models behave as we expect. Indeed, we observed that the baseline model produces quite a lot of insertions, while the one trained with Disfluencies Train and partial words removed does not.\ Example transcripts are depicted in \cref{fig:ex1}.

\begin{figure*}[t]
  		\centering
  		\includegraphics[width=1\linewidth, trim={0cm 13.7cm 2cm 0cm},clip]{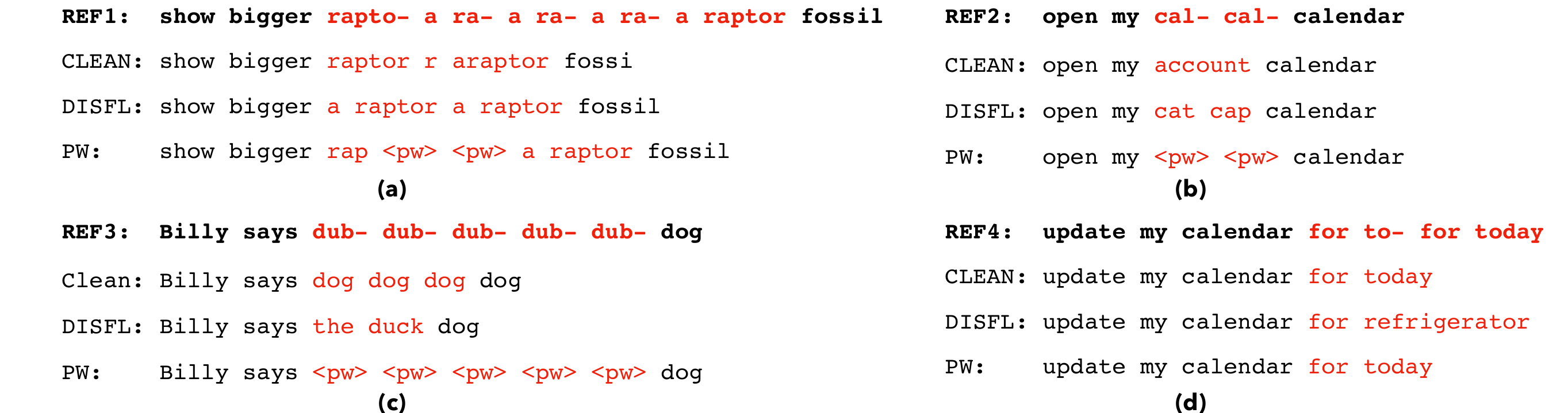}
  	\caption{Example recognition results. \textit{REF} denotes the reference transcription, \textit{CLEAN} was produced by the baseline model, \textit{DISFL} -- by the model trained on Ordinary Train merged with Disfluencies Train and with partial words removed from training transcripts, \textit{PW} -- same as \textit{DISFL} but with partial words replaced by a tag.}
  	\label{fig:ex2all}
\end{figure*}

The additional examples are presented in \cref{fig:ex2all} including those derived with the model trained with the replacement of partial words by $\langle\text{pw}\rangle$ in transcripts for training.\ As one would expect, the model trained with $\langle\text{pw}\rangle$ produces reasonable 'alignments' in some cases capturing the amount of partial words uttered as in \cref{fig:ex2all}b-c, while not so reasonable in the others (\cref{fig:ex2all}a). In some cases (\cref{fig:ex2all}d) this model produces a better result than the one with partial words removed without outputting the $\langle\text{pw}\rangle$ tag.\

One can speculate that mapping all partial words to a single tag allows the model to capture acoustic and lingustic patterns associated with partial word appearance and to preserve the integrity of the 'normal' speech patterns which would not happen if partial words were removed from the transcripts. 

\begin{table*}[t]	
	\setlength{\tabcolsep}{0.38em}\renewcommand{\arraystretch}{1.1}  
	\centering
	\caption{Evaluation results on different test sets depending on partial words handling in transcripts. Model-specific post-processing was applied before evaluation to get rid of partial words, or $\langle\text{pw}\rangle$ tag. Partial words were removed from reference transcriptions as well. \textit{NWER} column contains the corresponding model WER divided by WER of the baseline model on Ordinary Test.  \textit{WERR} (\%) is $100*(y-x)/y$ where $x$ is the corresponding model WER and $y$ is WER of the baseline model on the same test set. \textit{S, I, D} columns contain shares (\%) of substitutions, insertions, deletions in the observed WER.}
	\label{tab:res-asr}
	\begin{tabular}{c l l c c >{\small} c >{\small} c  >{\small} c c c >{\small} c >{\small} c  >{\small} c c c >{\small} c >{\small} c >{\small} c}
		\hline			
\multirow{2}{*}{\textbf{\#}} & \multirow{1}{*}{\textbf{Ordinary}} &\multirow{1}{*}{\textbf{Partial }} & \multicolumn{5}{c}{\textbf{Ordinary Test}} & \multicolumn{5}{c}{\textbf{Disfluencies Test}}& \multicolumn{5}{c}{\textbf{Stutter Test}}\\ \cline{4-18}
& \textbf{Train and...} & \textbf{words are ...}&\footnotesize NWER &\footnotesize WERR &  S & I &  D &\footnotesize NWER &\footnotesize WERR &  S & I &  D &\footnotesize NWER &\footnotesize WERR &S & I & D\\ \hline \hline
1 & \mbox{-} & absent & 1 & 0.0 & 60 & 18 &  23 & 3.02 & 0.0 & 29 & 57 & 13 & 2.35 & 0.0 & 34 & 49 & 17 \\ \hline
2 & \multirow{4}{*}{}& deleted &  1 & 0.2 & 59 & 18 & 23 & 2.38 & 21.0& 33 & 47 & 20 & 2.02 & 14.0 & 40 & 39 & 21 \\
3 & Disfluencies & replaced by $\langle\text{pw}\rangle$ & 1 & 0.5 & 59 & 17 & 24 & \textbf{2.34} & \textbf{22.4} & 33 & 45 & 22 & \textbf{1.93} & \textbf{17.9} & 39 & 34 & 29 \\
4 & Train & appended by $\langle\text{pw}\rangle$ & 1 & 0.2 & 59 & 18 & 23 & 2.6 & 13.7 & 31 & 51 & 18 & 2.33 & 0.9 & 37 & 45 & 18 \\
5 & &  \multirow{1}{*}{\footnotesize replaced by 1st letter \&$\langle\text{pw}\rangle$ } & 1 & 0.4 & 59 & 17 & 24 & 2.49 & 17.4 & 31 & 49 & 20 & 2.1 & 10.5 & 36 & 40 & 23 \\ \hline
\end{tabular}
\end{table*} 

\subsection{Word Error Rates}

In \cref{tab:res-asr} one can find word error rates for different models trained.\ The Ordinary Test results are provided to make sure that while improving on data with disfluencies we don't have degradation on this dataset.\ As expected, the error on the test sets with disfluencies is more than $2$ times higher than on Ordinary Test and the baseline model produced the highest number of insertions on all three test sets.\ By adding Disfluencies Train with removed partial words we achieved $21\%$ and $14\%$ relative WER reduction on the the test sets with disfluencies in comparison to the baseline. If partial words are replaced by a tag, we see a modest additional reduction by $1.7\%$ and $3.9\%$. When we try to preserve all or some characters of a disfluency, the reduction is much smaller (lines $4$ and $5$ in \cref{tab:res-asr}).\ It may seem surprising that WER on Stutter Test is significantly lower, than on Disfluencies Test.\ This is explained by the nature of the data: the former was recorded in a quiet room with a limited set of popular prompts, while the latter contains challenging field data.\ 

In order to verify reproducibility of the observed effects and to investigate how much data with disfluencies is actually needed to improve WER for speakers with stuttering, we conducted additional experiments. We used only the training sets where partial words were removed or replaced by $\langle\text{pw}\rangle$ because the corresponding models demonstrated much lower WER values than the others. Trainings with $3$ different seeds were performed for each dataset configuration, then each model was evaluated and the  WER numbers were averaged between the runs. The results are summarized in \cref{tab:res-asr-data}.\ The models corresponding to lines $4$ and $5$ give lower WER on the Disfluencies and Stutter datasets than the model number $6$, which confirms benefits from using a tag to denote all partial words in transcripts.\ Another observation is that as the amount of data with disfluencies increased, there was a gradual decrease in WER on Disfluencies and Stutter datasets with saturation more pronounced for the latter one. Probably in order to further improve accuracy one needs to take into account additional aspects associated with stuttering which we don't pay attention to.

We emphasise that the natural frequency of partial words appearing in the dataset available to us is rather low due to the heavy head of the requests distribution, so even $1/4$ of the Disfluencies Train should be considered as oversampling (it constitutes about 0.5\% of Ordinary Train size).
	
\section{Conclusions}
\label{section:conclusions}
In this work we showed that RNN-T based speech recognition models tend to produce insertions when presented with speech containing partial words if data with such words were not included in the training set. This contributes to low recognition accuracy for speakers with a stuttering disorder. Adding the data with partial words to the training set and increasing their relative share leads to significant WER reduction on the test sets with disfluencies without accuracy degradation on the average data. Replacing partial words in transcripts with a tag for training allows to reach even lower WER. Relative to the baseline the best model configuration allowed to achieve $22\%$ reduction on the test with disfluencies and $16\%$  on the test containing stuttering speech.

\begin{table}[H]	
	\setlength{\tabcolsep}{0.6em}\renewcommand{\arraystretch}{1.0}  
	\centering
	\caption{WER reduction relative to the baseline model (\%) depending on the fraction of the Disfluencies Train dataset used for training.}
	\label{tab:res-asr-data}
	\begin{tabular}{cllcccc} 
		\hline			
\multirow{2}{*}{\textbf{\#}} & \small  \multirow{1}{*}{\textbf{Ordinary}} & \small  \multirow{1}{*}{\textbf{Partial}} & \multicolumn{1}{c}{ \small \textbf{Ordin.}} & \multicolumn{1}{c}{\small \textbf{Disfl.}} & \multicolumn{1}{c}{\small \textbf{Stutter}} \\
& \textbf{\small Train and ...} & \small  \textbf{words are ...}& \small  \textbf{Test} & \small  \textbf{Test} & \small  \textbf{Test} \\ \hline \hline
1 & $\mbox{-}$ &  absent & 0.0 & 0.0 & 0.0 \\ \hline
2 & 1/10 Disfl. & \multirow{4}{*}{}& 0.3 & 8.7 & 5.8 \\
3 & 1/4 Dislf. & replaced by & 0.2 & 13.9 & 6.6 \\
4 & 1/2 Disfl. & $\langle\text{pw}\rangle$ & 0.0 & 18.3 & 14.5 \\
5 & Full Disfl. & & 0.0 & \textbf{22.5} & \textbf{16.4} \\ \hline
6 & Full Disfl. & deleted & -0.1 & 19.1 & 13.1 \\ \hline
\end{tabular}
\end{table} 

\section{Future work}
We see two directions for the future work which benefit each other. The first is increasing ASR robustness to disfluencies occurring in fluent speech by using data augmentation, semi-supervised learning approaches.\ The second is pushing the boundary of what an ASR system can do out-of-the-box for speakers with speech less fluent due to stutter, age or other factors.

\vfill\pagebreak

\bibliographystyle{IEEEbib}
\bibliography{strings,refs}

\begin{thebibliography}{10}

\bibitem{rnnt}
Alex Graves,
\newblock ``Sequence transduction with recurrent neural networks,''
\newblock {\em arXiv:1211.3711}, 2012.

\bibitem{preliminar}
Elizabeth~Ellen Shriberg,
\newblock {\em Preliminaries to a theory of speech disfluencies},
\newblock Ph.D. thesis, Citeseer, 1994.

\bibitem{heeman1999speech}
Peter~A Heeman and James Allen,
\newblock ``Speech repains, intonational phrases, and discourse markers:
  modeling speakers’ utterances in spoken dialogue,''
\newblock {\em Computational Linguistics}, vol. 25, no. 4, pp. 527--572, 1999.

\bibitem{johnson2004tag}
Mark Johnson and Eugene Charniak,
\newblock ``A tag-based noisy-channel model of speech repairs,''
\newblock in {\em Proceedings of the 42nd Annual Meeting of the Association for
  Computational Linguistics (ACL-04)}, 2004, pp. 33--39.

\bibitem{honal2003correction}
Matthias Honal and Tanja Schultz,
\newblock ``Correction of disfluencies in spontaneous speech using a
  noisy-channel approach,''
\newblock in {\em Eighth European Conference on Speech Communication and
  Technology}, 2003.

\bibitem{shriberg1999phonetic}
Elizabeth~E Shriberg,
\newblock ``Phonetic consequences of speech disfluency,''
\newblock Tech. {R}ep., SRI INTERNATIONAL MENLO PARK CA, 1999.

\bibitem{liu2005comparing}
Yang Liu, Elizabeth Shriberg, Andreas Stolcke, and Mary Harper,
\newblock ``Comparing hmm, maximum entropy, and conditional random fields for
  disfluency detection,''
\newblock in {\em Ninth European Conference on Speech Communication and
  Technology}, 2005.

\bibitem{ferguson2015disfluency}
James Ferguson, Greg Durrett, and Dan Klein,
\newblock ``Disfluency detection with a semi-markov model and prosodic
  features,''
\newblock in {\em Proceedings of the 2015 Conference of the North American
  Chapter of the Association for Computational Linguistics: Human Language
  Technologies}, 2015, pp. 257--262.

\bibitem{zayats2016disfluency}
Vicky Zayats, Mari Ostendorf, and Hannaneh Hajishirzi,
\newblock ``Disfluency detection using a bidirectional lstm,''
\newblock {\em arXiv:1604.03209}, 2016.

\bibitem{alharbi2020sequence}
Sadeen Alharbi, Madina Hasan, Anthony~JH Simons, Shelagh Brumfitt, and Phil
  Green,
\newblock ``Sequence labeling to detect stuttering events in read speech,''
\newblock {\em Computer Speech \& Language}, vol. 62, pp. 101052, 2020.

\bibitem{bach2019noisy}
Nguyen Bach and Fei Huang,
\newblock ``Noisy bilstm-based models for disfluency detection.,''
\newblock in {\em Proc. Interspeech}, 2019, pp. 4230--4234.

\bibitem{kourkounakis2020detecting}
Tedd Kourkounakis, Amirhossein Hajavi, and Ali Etemad,
\newblock ``Detecting multiple speech disfluencies using a deep residual
  network with bidirectional long short-term memory,''
\newblock in {\em Proc. IEEE Intern. Conf. on Acoustics, Speech and Signal
  Process. (ICASSP)}, 2020, pp. 6089--6093.

\bibitem{shor2019personalizing}
Joel Shor, Dotan Emanuel, Oran Lang, Omry Tuval, Michael Brenner, Julie
  Cattiau, Fernando Vieira, Maeve McNally, Taylor Charbonneau, Melissa
  Nollstadt, et~al.,
\newblock ``Personalizing asr for dysarthric and accented speech with limited
  data,''
\newblock {\em arXiv:1907.13511}, 2019.

\bibitem{graves2013speech}
Alex Graves, Abdel-rahman Mohamed, and Geoffrey Hinton,
\newblock ``Speech recognition with deep recurrent neural networks,''
\newblock in {\em 2013 IEEE international conference on acoustics, speech and
  signal processing}. IEEE, 2013, pp. 6645--6649.

\bibitem{hochreiter1997long}
Sepp Hochreiter and J{\"u}rgen Schmidhuber,
\newblock ``Long short-term memory,''
\newblock {\em Neural computation}, vol. 9, no. 8, pp. 1735--1780, 1997.

\bibitem{kudo2018sentencepiece}
Taku Kudo and John Richardson,
\newblock ``Sentencepiece: A simple and language independent subword tokenizer
  and detokenizer for neural text processing,''
\newblock {\em arXiv preprint arXiv:1808.06226}, 2018.

\bibitem{kingma2014adam}
Diederik~P Kingma and Jimmy Ba,
\newblock ``Adam: A method for stochastic optimization,''
\newblock {\em arXiv preprint arXiv:1412.6980}, 2014.

\bibitem{park2019specaugment}
Daniel~S Park, William Chan, Yu~Zhang, Chung-Cheng Chiu, Barret Zoph, Ekin~D
  Cubuk, and Quoc~V Le,
\newblock ``Specaugment: A simple data augmentation method for automatic speech
  recognition,''
\newblock {\em arXiv preprint arXiv:1904.08779}, 2019.

\end{thebibliography}

\end{document}